%% file: arxiv.tex
\documentclass[10pt,twocolumn,letterpaper]{article}

\usepackage[pagenumbers]{cvpr} 

\usepackage{graphicx}
\usepackage{amsmath}
\usepackage{amssymb}
\usepackage{booktabs}

\usepackage{algorithm}
\usepackage{algpseudocode}

\usepackage[pagebackref,breaklinks,colorlinks]{hyperref}

\usepackage[capitalize]{cleveref}
\crefname{section}{Sec.}{Secs.}
\Crefname{section}{Section}{Sections}
\Crefname{table}{Table}{Tables}
\crefname{table}{Tab.}{Tabs.}

\def\cvprPaperID{12507} 
\def\confName{CVPR}
\def\confYear{2023}

\usepackage[rgb,dvipsnames]{xcolor}
\definecolor{OliveGreen}{rgb}{0,0.6,0}
\definecolor{Burgundy}{RGB}{144,0,32}

\newcommand{\authorname}[1]{\fontsize{11}{11}\selectfont #1}
\makeatletter
\def\@fnsymbol#1{\ensuremath{\ifcase#1\or \dagger\or \ddagger\or
   \mathsection\or \mathparagraph\or \|\or **\or \dagger\dagger
   \or \ddagger\ddagger \else\@ctrerr\fi}}
\makeatother

\begin{document}
\title{Text-Conditional Contextualized Avatars For Zero-Shot Personalization}

\author{
\authorname{Samaneh Azadi}$^{*}$ \\
\authorname{Guan Pang} 
\and
\authorname{Thomas Hayes}$^{*}$ \\
\authorname{Devi Parikh} \\ \\
Meta AI \\
{\tt\small \{azadis, thayes427\}@meta.com} \\
\and
\authorname{Akbar Shah} \\
\authorname{Sonal Gupta} \\
}

\makeatletter
\let\@oldmaketitle\@maketitle
\renewcommand{\@maketitle}{\@oldmaketitle
\centering
\includegraphics[width=0.9\linewidth]{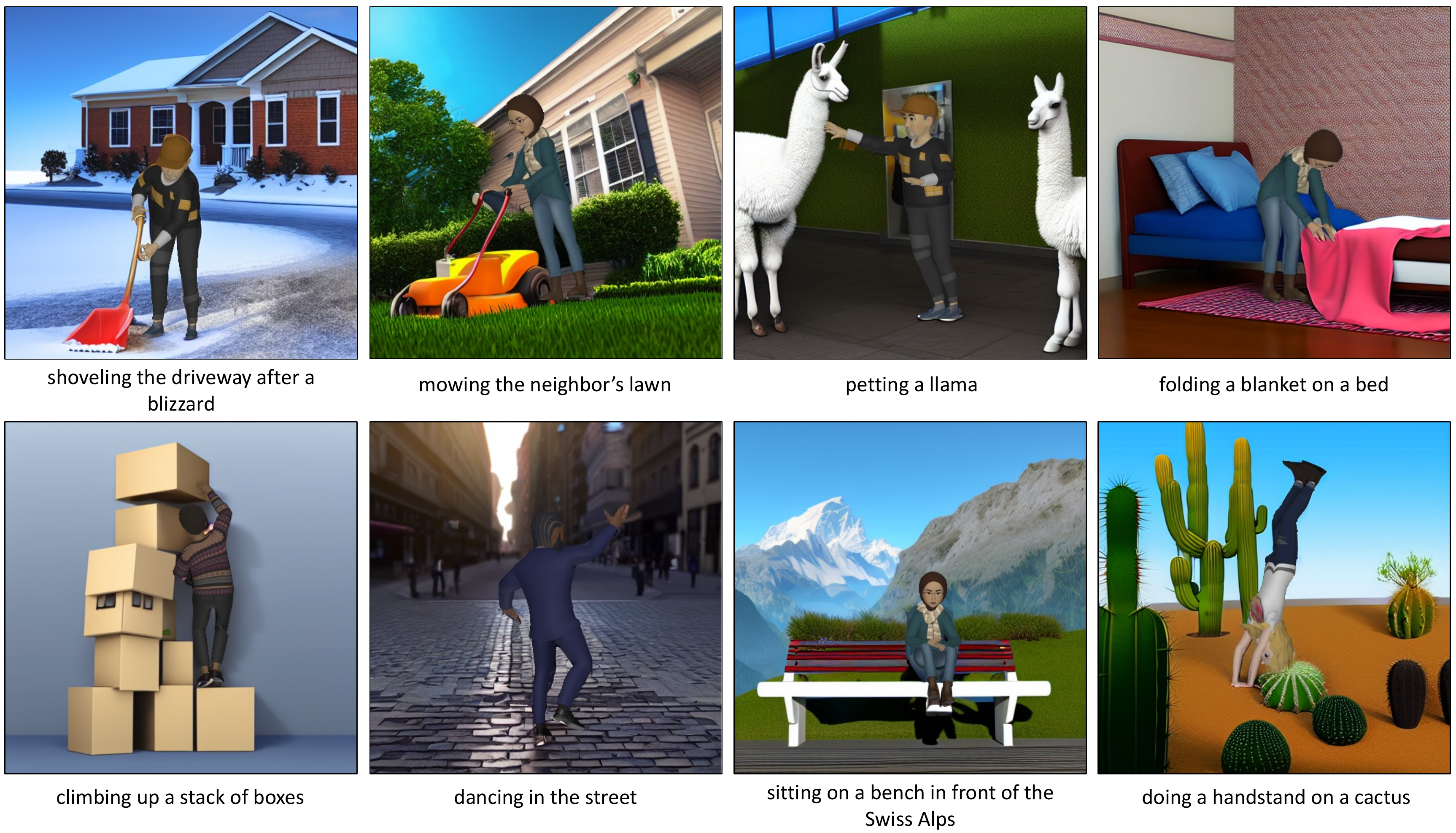}
\captionof{figure}{Samples of images generated by our proposed approach Personalized Avatar Scene (PAS). Each caption is prefixed by ``A person (is) ''.}
\label{fig:teaser}
\bigskip\bigskip}
\makeatother

\maketitle
\begin{abstract}
Recent large-scale text-to-image generation models have made significant improvements in the quality, realism, and diversity of the synthesized images and enable users to control the created content through language. However, the personalization aspect of these generative models is still challenging and under-explored. In this work, we propose a pipeline that enables personalization of image generation with avatars capturing a user’s identity in a delightful way. Our pipeline is zero-shot, avatar texture and style agnostic, and does not require training on the avatar at all - it is scalable to millions of users who can generate a scene with their avatar. To render the avatar in a pose faithful to the given text prompt, we propose a novel text-to-3D pose diffusion model trained on a curated large-scale dataset of in-the-wild human poses improving the performance of the SOTA text-to-motion models significantly. We show, for the first time, how to leverage large-scale image datasets to learn human 3D pose parameters and overcome the limitations of motion capture datasets. 
\end{abstract}

\def\thefootnote{*}\footnotetext{ Denotes equal contribution.}\def\thefootnote{\arabic{footnote}}

\section{Introduction}
Millions of people are connected with their family and friends through social media and chats where self-expression is not only possible through language but more efficiently through visual features such as stickers and avatars. Avatars are a befitting method for personalization as they enable users to inject their identity into an expressive virtual self, while mitigating deepfake and privacy concerns. With current technology, however, one's expression is limited to a set of predefined stickers and avatars created by designers. Imagine you have the ability to generate images of your avatar in any context you desire — you can travel to your favorite city in the world, walk on your favorite beach, or even float in outer space. 

Recent large-scale text-to-image generation models~\cite{ramesh2022dalle2} have made significant improvements in the quality, realism, and diversity of synthesized images and enable users to control content generation through language. However, the personalization aspect of these generative models is still challenging and under-explored. Dreambooth\cite{ruiz2022dreambooth} enabled personalization of text-to-image generation models by fine-tuning on a few instances of a subject. While it can produce delightful results, it is not scalable to millions of users since fine-tuning is required for each new subject. Furthermore, it can sometimes generate images unfaithful to the subject's identity. Other image personalization methods suffer from similar limitations: poor faithfulness to user's identity, not zero-shot, and/or lacking open-world text understanding.

\begin{figure*}
  \centering \includegraphics[width=\textwidth]{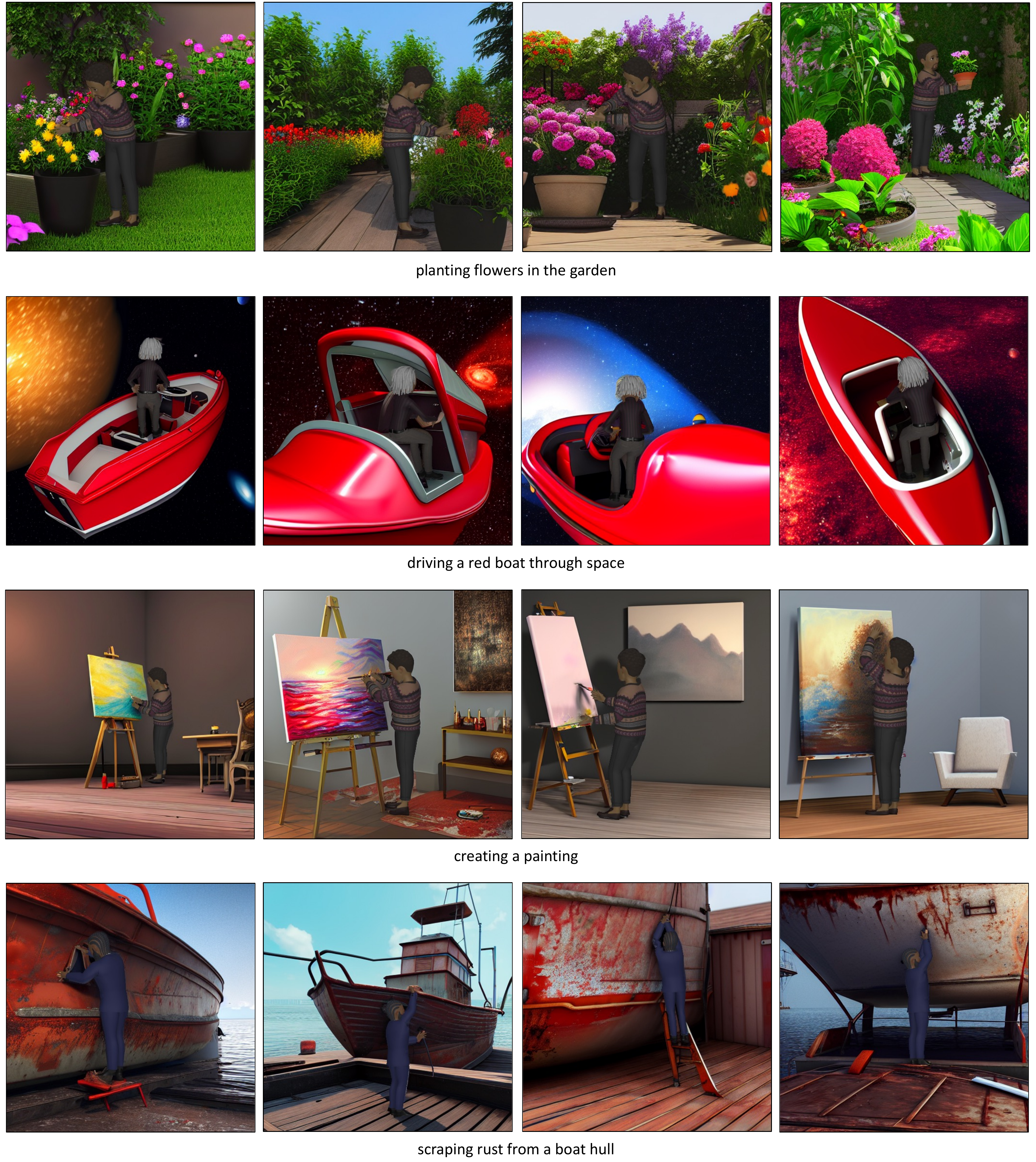}
  \caption{Multiple samples for the same prompt showing diversity in pose and image generation. Each caption is prefixed by ``A person (is) ''.}
  \label{fig:img_sample_diversity}
\end{figure*}

In this work, we propose a method to overcome these limitations, Personalized Avatar Scene (PAS), which enables zero-shot personalization of text-to-image generation with avatars (sample images in Fig.~\ref{fig:teaser}, Fig.~\ref{fig:img_sample_diversity}). Our method is agnostic to the avatar style, requires no fine-tuning on the user's avatar (or even training on avatars at all), and remains faithful to the avatar appearance, making it truly scalable to millions of users. 

Our pipeline involves a three step approach: (1) generate human body pose from text, (2) render a user's avatar in the generated pose, (3) condition an image generation model on the input text and the rendered avatar to generate an image of the avatar contextualized in a scene. Re-posing and rendering a user's avatar enables us to maintain strict faithfulness to their appearance. Zero-shot capabilities come from the fact that our image generation model learns to faithfully place the rendered avatar in to scene, regardless of its style or degree of photo-realism. 

We represent the avatar body pose via 3D SMPL parameters and learn text-to-3D pose generation with a Transformer-based diffusion model. We train our pose generation model on a large-scale dataset of text-pose pairs constructed by extracting 3D pseudo-pose SMPL annotations from image-text datasets filtered for images containing humans. This large-scale Image Text Pseudo-Pose (ITPP) dataset helps to overcome the limited diversity of existing motion capture datasets, in terms of both pose and text diversity, and demonstrates dataset preparation required for human motion generation can be significantly simplified. Our 3D human pose generation model reaches state-of-the-art performance. 

By representing avatar body poses via 3D SMPL parameters, our method is easily adaptable to a number of rendering engines to personalize with the user's avatar appearance.

After rendering the avatar in a pose that aligns with the text prompt, we contextualize the avatar in a scene by leveraging large-scale text-to-image generation models to outpaint from the avatar. We use a pre-trained a text-to-image model that has been trained on the large-scale image-text datasets and fine-tune it on the corresponding human masks and pseudo-pose annotations to generate images more faithful to the avatar appearance. To summarize, our main contributions are:
\begin{itemize}
    \item We present Personalized Avatar Scene (PAS) -- a scalable method for zero-shot personalization of image generation with a user's avatar.
    \item We show, for the first time, how to leverage large-scale image datasets to learn in-the-wild human poses. Our text-to-3D pose model significantly improves on prior state-of-the-art models.
    \item We present a method for zero-shot personalized image generation which improves faithfulness to the user's identity relative to baselines by injecting human body priors. 
\end{itemize}

\section{Related Works}

\subsection{Human pose and motion synthesis}
Prior works in human pose and motion synthesis have explored generative models either unconditionally~\cite{yan2019, zhao2020}, or conditioned on various input signals such as a prior motion~\cite{julieta2017, ruiz2019}, an action class~\cite{guo2020a2m, petrovich2021actor, cervantes2022}, or music~\cite{lee2019, li2021}. Using text descriptions as a guidance in pose and motion synthesis has been a more recent research direction where many existing works use 2D keypoints as pose representations. \cite{zhou2019pose} select a base pose from 8 clusters based on an input text fed to a GAN model to generate a human image. \cite{zhang2021} use GAN to generate a set of heatmaps for body keypoints conditioned on the text input. \cite{roy2022} propose a coarse-to-fine approach where a refinement stage is introduced on top of the initial coarse estimate of the keypoint heatmaps. Different from the above methods, \cite{briq2021} use SMPL to represent a 3D body pose, generated by an LSTM GAN from a text input. Here, we similarly choose the 3D SMPL pose representation to facilitate more expressive body movements, utilize any existing 3D rendering engines, and enable a high-quality zero-shot personalization.

Text-guided motion generation can be regarded as an extension to pose generation, where motion is a sequence of poses. Many existing work approach this problem by learning to align text and pose or motion embeddings in the feature space. JL2P~\cite{ahuja2019jl2p} proposes to learn the joint embedding of text and pose using an autoencoder with curriculum learning. \cite{ghosh2021} propose a two-stream model to encode upper and lower body motions separately. AvatarCLIP~\cite{hong2022avatarclip} uses a pre-trained VPoser model to generate candidate poses, which are then used to optimize a motion VAE. MotionCLIP~\cite{tevet2022motionclip} trains an auto-encoder while simultaneously reconstructing motion and aligning the motion manifold with CLIP's latent space. TEMOS~\cite{petrovich2022temos} trains a joint latent space through separate text and motion transformer encoders, allowing non-deterministic motion sampling. T2M~\cite{guo2022t2m} also uses a VAE model, but encodes motion as snippets and introduces an extra sampling for motion length conditioned on the input text.

Leveraging recent advancements in diffusion models, Motion Diffusion Model (MDM)~\cite{tevet2022mdm} and MotionDiffuse~\cite{zhang2022md} both adopt a diffusion process from text to motion with a transformer, which can significantly improve the diversity of synthesized samples. Both these methods are trained on HumanML3D~\cite{guo2022t2m} which partially limits their choice of representation to stick figures, while our curated large-scale dataset provides millions of SMPL pose labels. We concurrently developed our text-to-3D pose diffusion model with a more efficient choice of pose representations as discussed in sec~\ref{sec:vposer}, whereas MDM supports converting poses from stick figures to SMPL bodies only through an optimization SIMPLIFY procedure that often results in unrealistic body poses. 

\begin{figure*}
  \centering \includegraphics[width=\textwidth]{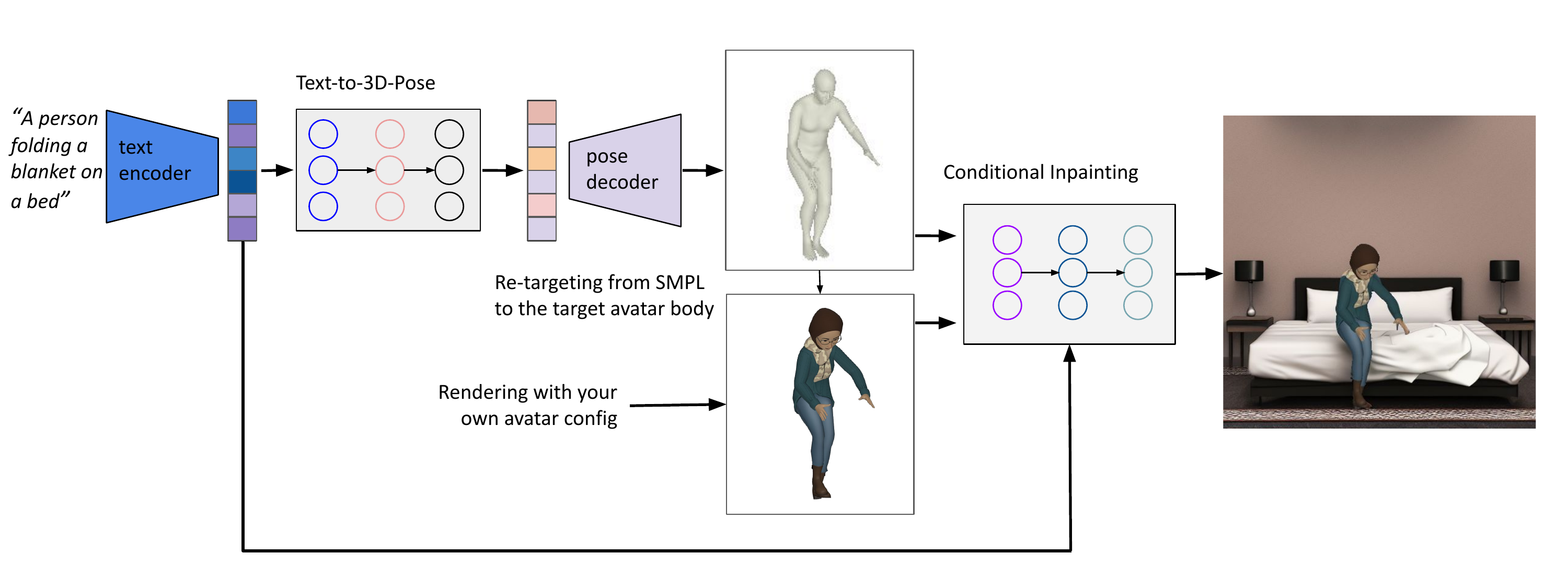}
  \caption{Personalized Avatar Scene (PAS): We generate 3D SMPL body poses using a diffusion based transformer model and leverage a pre-trained VPoser either for pose regularization or decoding. The generated pose is re-targeted to the avatar body enabling every user to render their own avatar in the target generated pose. Finally, we generate an avatar scene using a fine-tuned text-to-image model conditioned on the rendered avatar and the text prompt.}
  \label{fig:postreg}
\end{figure*}

\subsection{Text-Conditional Image Synthesis}
Text-to-image generation studies the task of synthesizing images from natural language descriptions. Early works \cite{reed16GANText} used conditional Generative Adversarial Networks \cite{goodfellow2014, Mirza2014ConditionalGA} in constrained domains (e.g., flowers and birds). GANs have been scaled up \cite{stylegan-xl2022, stylegan-T2023} using a progressive growing training strategy, which has led to success in more diverse domains.

In parallel with the works improving GAN-based methods, the field has found success by treating text-to-image generation as a sequence modeling problem. DALL-E\cite{DALLE2021} employed a two stage approach, whereby in the first stage a VQ-VAE is learned to compress images into discrete tokens, and in the second stage a Transformer decoder is trained to model text-to-image generation as a sequence-to-sequence problem. Make-A-Scene~\cite{make_a_scene} improved on this approach by implicitly modeling intermediate semantic maps and by adding human priors to improve the tokenization reconstruction quality. Recently, Parti~\cite{Parti} employed a Transformer-based image tokenizer, ViT-VQGAN, to further improve reconstruction quality, and scaled an encoder-decoder Transformer up to 20B parameters.

Lately, the field has seen tremendous progress in terms of sample fidelity and alignment with text using Denoising Diffusion Probablistic Models (DDPMs)~\cite{ddpms}, a class of latent variable models which have shown superior image generation quality compared to GANs ~\cite{diff_beat_gans}. Several follow-up works~\cite{GLIDE2022, Ramesh2022HierarchicalTI, imagen} leverage diffusion models in the pixel space for text-to-image generation and achieve high quality results. These approaches opt for a cascade of diffusion models which progressively upsample the spatial resolution.
Our model is fine-tuned based on a pre-trained diffusion model operating in an auto-encoder latent space to generate an image conditioned on a full body avatar in a given pose.

\subsection{Personalization of Image Generation}
Personalization of generative models has been a challenging problem. One way to approach personalization of image generation is to consider it as novel synthesis of a given subject. CompositionalGAN~\cite{azadi2020compositional} proposed a model to compose two given objects in a novel and realistic way and was focused on the relation between two specific domains and objects while keeping their identity unchanged. Dreambooth~\cite{ruiz2022dreambooth} proposes a few-shot model that generates novel samples of a subject controlled by a text prompt. However, the primary limitation of this approach is that this model must be fine-tuned on each subject and requires 3-5 instances per subject. Fine-tuning large-scale text-to-image models is rather inaccessible for much of the community, is sensitive to hyper-parameter tuning, and is not scalable to millions of users.

Image editing has been another approach towards personalization through local or global edits to an image or an object~\cite{hertz2022prompt, nichol2021glide}. Imagic~\cite{kawar2022imagic} enables the ability to apply complex text-guided semantic edits to a real image such as editing the posture and composition of one or multiple objects inside an image, while preserving their original characteristics. Semantic diffusion guidance proposed in~\cite{liu2023more} enables generating variations of an input image while preserving its content. 
 
Personalization has been also studied in novel view and garment synthesis of humans while preserving their identity~\cite{sarkar2021style} with applications in human appearance transfer, virtual try-on, and motion imitation. Some works on neural rendering~\cite{liu2020neural, meshry2019neural, kappel2021high} generate a 2D/3D skeleton or a heatmap as well as a coarse neural texture and then use image-to-image translation techniques to synthesize the appearance in higher resolution. Other works study human pose transfer as an image-to-image mapping problem given a reference image of a target person. In these methods body pose is represented as a rendering of a skeleton~\cite{zhu2019progressive, chan2019everybody}, dense mesh~\cite{grigorev2019coordinate, kappel2021high}, or joint position heatmaps~\cite{aberman2019deep, ma2018disentangled, ma2017pose}. However, most of these learning based approaches have limits in the visual quality and add artifacts which can change the identity of the person.

In this work, we introduce personalization of image generation via avatars and enable their interaction with different objects and scenes controlled by a text prompt. We represent body pose through 3D SMPL parameters that make them compatible with any rendering service to generate an avatar's appearance in high quality and preserve its identity. Our model can serve all people regardless of avatar texture, style, or appearance—it is zero-shot. No model retraining is required when avatar style or appearance changes.

\section{Personalized Avatar Scene (PAS)}
\begin{figure}
  \centering \includegraphics[width=0.5\textwidth]{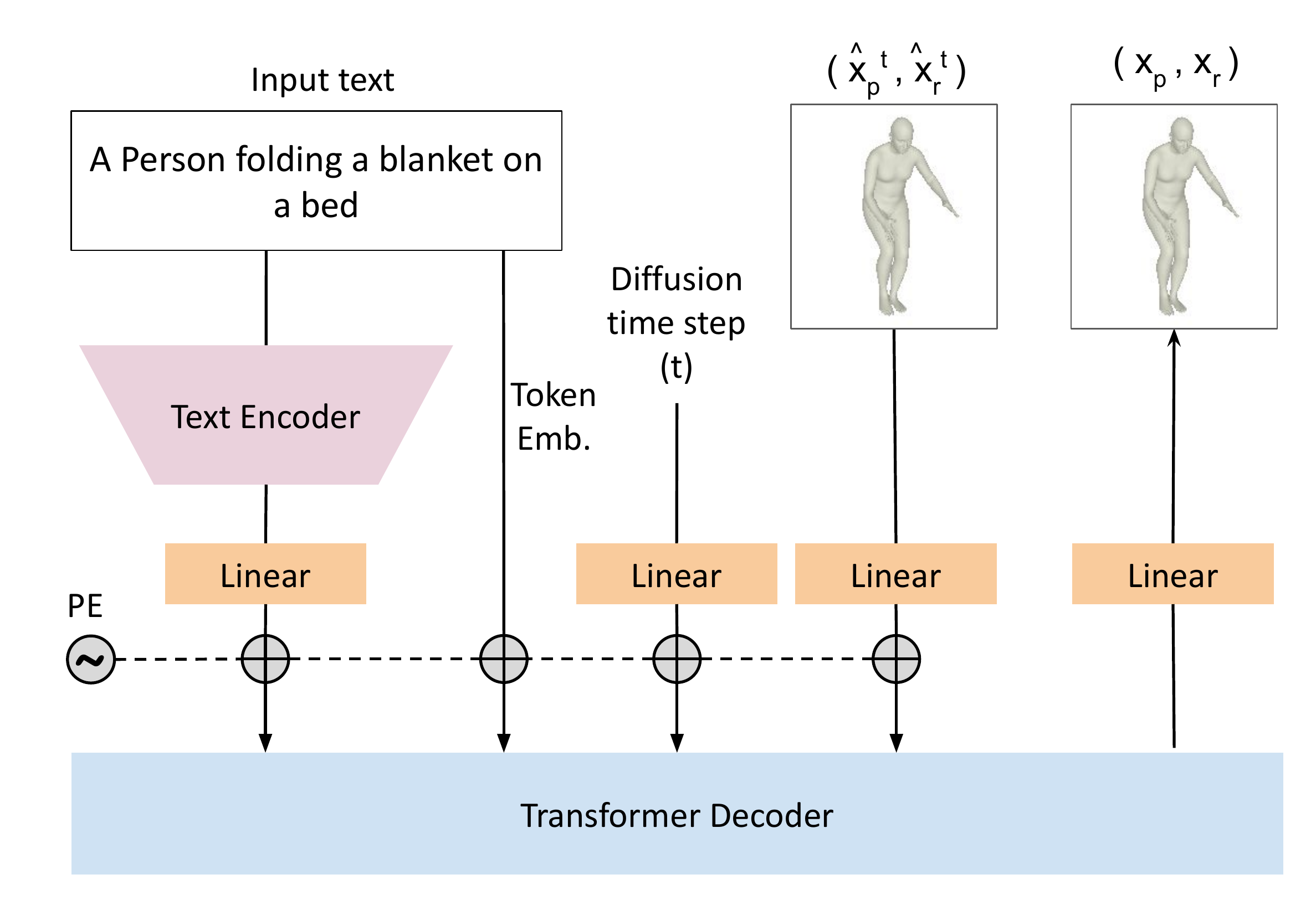}
  \caption{Our transformer based Text-to-3D pose diffusion model at time step t. The input sequence includes CLIP text embedding, tokens embedding, diffusion timestep, and noised pose and root orient representations  $(\hat{x}_p,\hat{x}_r)$ all projected to the transformer dimension. A positional embedding is added to each token in the above sequence. The un-noised pose and root orient representations are predicted at each timestep during training.}
  \label{fig:posegen}
\end{figure}

We represent an avatar body pose via 3D SMPL parameters, learn to predict them from text via a Transformer-based diffusion model, and then retarget to the avatar body to be rendered by our rendering tool. We train our pose generation model on a large-scale dataset of human poses, as described in section~\ref{sec:data}. Our image generation model is conditioned on the rendered avatar in the target pose as well as a text prompt describing the avatar's action and the scene. The schematic of our model is illustrated in Figure~\ref{fig:postreg}.

\subsection{Background: Variational Human Pose Prior}
\label{sec:vposer}
Since 3D pose of human body is complex and high dimensional, we use a pre-trained VPoser~\cite{SMPL-X:2019} as a Variational Autoencoder to efficiently model the distribution of pose priors instead of learning to generate all 3D joint rotations. VPoser learns a latent representation of human pose and regularizes the distribution of the latent code to be a normal distribution. The VPoser has been pre-trained on a large dataset of SMPL pose representations using the Kullback-Leibler and reconstruction losses (following the common VAE formulation), in addition to losses which encourage encoding of valid rotation matrices. The architecture consists of a symmetric encoder-decoder, each with two dense layers, and a 32 dimensional vector as the bottleneck latent space. 

This continuous embedding space fits well in the the gradual noising and denoising diffusion process, while the widely used human pose representations such as quaternions and Euler angles are discontinuous~\cite{zhou2019continuity}.

For some experiments we alternatively train our model on the 6D continuous pose representations for the 21 SMPL joints and root orient. Since joint representations are unconstrained in this setting, unnatural body and hand poses can be expected. As a post-processing inference step after generating samples with this 6D pose representation, we pass poses to the VPoser encoder and decoder to regularize them in the natural distribution of human bodies.

\subsection{Large-Scale Image Text Pseudo-Pose (ITPP) Dataset}
\label{sec:data}
We have collected a dataset containing 35M pairs of human poses and their text descriptions from a few large-scale image-text datasets. We processed all images by running Detectron2 keypoint detector~\cite{wu2019detectron2} to find images with a single human, then extracted their 3D pseudo-pose SMPL annotations using a pre-trained PyMAF-X model~\cite{pymafx}. This large-scale data overcomes the limitations of the existing mocap datasets by providing a wide variety of human poses and a huge number of (text, 3D pose) sample pairs. Due to legal concerns, we do not extract any face or hand expressions from the aforementioned datasets.

\subsection{Text-to-3D Pose Synthesis}
We design a diffusion based generative model, $f_\theta$, that maps the input text encodings, $y$, from a pre-trained CLIP model to the concatenation of the continuous body pose representations and root orient, $(x_p, x_r)$. Our text-to-3D-pose model is based on a decoder-only Transformer with a causal attention mask operating on a sequence of tokenized captions and their CLIP text embeddings, the diffusion time-step embedding, the noised body pose and root orient representations, and two final pose and orientation queries to predict the unnoised pose and root orientation, respectively. The model architecture is illustrated in Figure~\ref{fig:posegen}. Similar to DALL-E 2 image prior~\cite{ramesh2022dalle2}, we train our model to directly predict the unnoised pose and root orientation using a mean-squared error loss as:

\begin{equation*}
\begin{split}
\mathcal{L}_{MSE} = \mathbb{E}_{t\sim [1,T], x_p\sim q_p, x_r\sim q_r} \parallel & f_\theta((x_p^{(t)}, x_r^{(t)}), t, y)  \\
& - (x_p, x_r)\parallel ^2
\end{split}
\end{equation*}

To improve sample quality at test time, we use classifier free guidance~\cite{cfg} by dropping the text conditioning 10$\%$ of the time during training. After training, we pass the body pose encodings generated by our diffusion model to the pre-trained VPoser decoder to generate the 3D SMPL body rotation vectors.

\subsection{Avatar Rendering}
To perform a zero-shot personalization given a generated 3D body pose, we first render an image of the posed avatar as a reference input. In order to personalize our image generation model with high-quality user avatars, we use an internal avatar representation and rendering engine. This demonstrates how our method is not restricted to the SMPL-based avatars.

Given the generated 3D pose in SMPL, we first perform a retargeting process to convert it to the internal avatar pose through an optimization process which matches the corresponding joints in position and orientation. Then, we take the converted pose and the personalized avatar config, including its shape and texture, to render an avatar image in the target pose, as shown in Fig.~\ref{fig:postreg}. Please note the retargeting and rendering steps do not need any training or prior knowledge of users' avatar configs, i.e., they are zero-shot. This setup can work with any other avatar representation as long as a retargeting solver from SMPL and rendering engine is in place.

\subsection{Text-to-Avatar Image via 3D Pose}

\subsubsection{High-Level Approach}
As opposed to methods which require fine-tuning for each subject of personalization, we desire a model which can perform zero-shot personalization to generate an image of a user's avatar interacting with a scene and objects corresponding to the input text. We can perform zero-shot personalization by re-framing it as an image outpainting task. That is, given a rendered image of a posed avatar and text which describes the scene and interactions, we generate a personalized image by outpainting from the avatar to fill in the rest of the scene and objects.

\subsubsection{Text-to-Image Priors}
Our model is a latent diffusion text-to-image generation model that has been pre-trained on large-scale image-text datasets enabling vast open-world visual textual understanding. Our model is built of a UNet in a learned latent space of an image autoencoder $z = \mathcal{E}(x)$ conditioned on the timestep $t$ and CLIP text encodings $y$ making it  more efficient than other large-scale text-to-image generation models.

\subsubsection{Personalized Image Generation Dataset}
To train this personalized diffusion model, we would ideally have a dataset of images of avatars interacting with open-world environments. However, no such dataset exists. Instead, we instead use our curated ITPP dataset of full body humans discussed in~\ref{sec:data}, and use panoptic segmentation~\cite{wu2019detectron2} to crop out the human for conditioning. So, at train time we condition the UNet on a cropped person, and at test time we condition on the image of the avatar rendered in the pose generated by the text-to-3D-Pose model.

\subsubsection{Model Architecture}
To improve the model’s faithfulness to the avatar’s appearance and reduce the domain gap between photo-realistic humans during train time and avatars during test time, we make two architecture modifications: (1) additionally condition the UNet on the gray body render of the 3D pose; (2) augment the personalization conditioning by downsampling the spatial dimensions and condition the UNet on the downsample factor. (1) Injects human body priors into the image generation model so it can generate more realistic avatar-object interactions. It also helps to bridge the gap between humans and avatars since gray body renders are the same for each.(2) helps the image generation model be less sensitive to the avatar's appearance when stylizing the image since downsampling removes some of the texture details in the conditioning.

Altogether, the full set of conditioning for the UNet is the text describing the scene and avatar-object interaction ($y$), the rendered RGBA image of the avatar in the target pose ($p$), the personalization downsample rate ($w$), and the timestep ($t$). We train the model via the loss:
\vspace{-3mm}
\begin{equation}
\begin{split}
\mathcal{L} = \mathbb{E}_{\mathcal{E}(x),y,p,w\sim [0,1], \epsilon\sim N(0,1),t\sim [1,T]} \Bigl[\parallel \epsilon \\ 
 - \epsilon_\theta(z_t, t, y, p, w)\parallel^2_2\Bigr] 
\end{split}
\end{equation}

To condition on the RGBA personalization image $p$, we use $\mathcal{E}(p)$ to encode the RGB channels to $z_p$ and concatenate $z_p$ onto $z_t$ along the channels dimension. We separately downsample the $\alpha$ channel to the spatial dimensions of $z$ (64x64) and concatenate along the channels dimension. To condition on the personalization augmentation downsample rate $w$, we encode it using a sinusoidal embedding, project to the dimensions of the CLIP text encodings, and concatenate with text encodings for cross-attention. Given that we formulate image personalization as an outpainting problem, it is natural to fine-tune from an inpainting/outpainting model which is already conditioned on $z_p$ and a corresponding mask.

\subsubsection{Post-Processing}
Maintaining strict faithfulness to the avatar's appearance is important to respect the user's identity. Because the diffusion autoencoder reconstruction is imperfect, we chose to paste the rendered RGBA image of the posed avatar back onto the generated image.

\section{Experiments}

In addition to our ITPP dataset, we also create a train and test set from the HumanML3D motion dataset~\cite{HumanML3d} to be used during training and test time, respectively. For each motion sequence, we render an avatar per frame and automatically split the corresponding long captions into smaller sentences based on their provided grammatical properties to make sure each text describes one frame. Then, we find top-10 and top-1 pairs of (rendered pose, caption) per motion sequence in terms of their image-text CLIP similarity score, resulting in 300K and 3250 samples in the train and test splits, respectively. 

We compare the performance of our pose generation model with the existing state-of-the-art text-to-human-motion generation models through human evaluation on 390 crowd-sourced prompts and automatic metrics on the HumanML3D test set. Given each text prompt, we generate five poses and render them through our rendering service. We then sort them based on the CLIP similarity scores between the rendered avatars as 2D images and the input text prompt to select the best generated sample. For each baseline, we generate a motion sequence given a text prompt and select the best representative frame based on their CLIP similarity scores with the text encoding. The results confirm the superiority of our pose generation system and its faithfulness to the input text. 

We also evaluate our personalized image generation model against a simple outpainting baseline on our crowd-sourced prompts and show that our fine-tuning and architecture modifications improve faithfulness to the avatar's appearance.

\subsection{Evaluation of Pose Generation}

\subsubsection{Human Evaluation Set and Metrics}
\label{sec:human_eval_set}
\begin{figure*}
  \centering \includegraphics[width=0.9\textwidth]{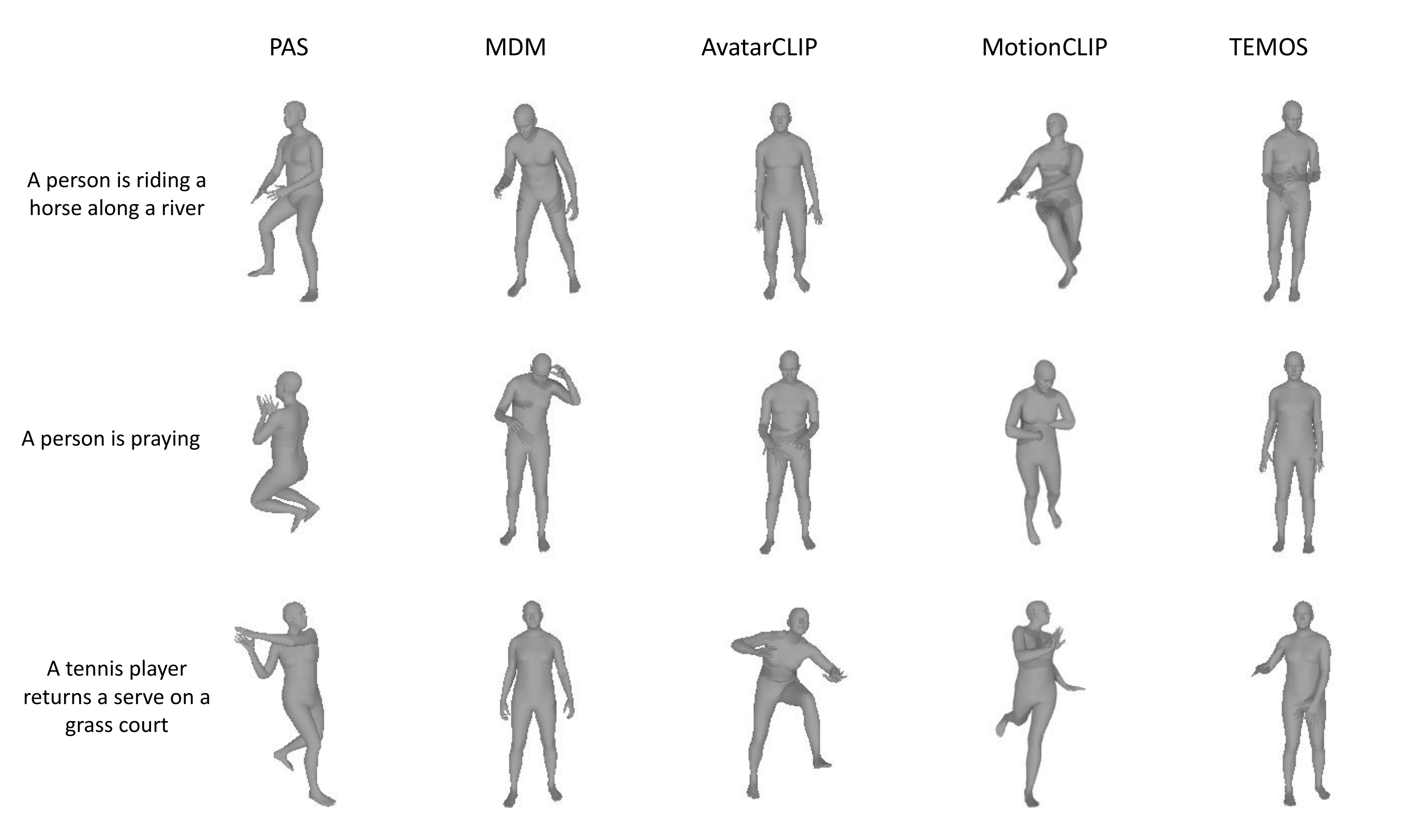}
  \caption{Qualitative comparison of our text-to-3D pose model (PAS) vs. other baselines.}
  \label{fig:ablation_pose}
\end{figure*}

We collect an evaluation set from Amazon Mechanical Turk (AMT) that consists of 390 prompts. We asked annotators for prompts that described an action along with some context of the scene. We filtered out prompts based on ethical concerns to remove any references to children or NSFW content. These prompts were selected without generating any poses or images for them, and were kept fixed for all of our evaluations. 
To evaluate faithfulness of the generated poses to the input texts, we show raters a text description and the corresponding rendered avatars from two models and ask them which output better matches the input text. For each comparison, we use the majority vote from $5$ different annotators as the final result, reported in Table~\ref{tab:pose_gen_eval}. This study confirms the superiority of our model and the impact of our large-scale ITPP dataset in human pose generation and overcoming the limitations of the MOCAP datasets. Qualitative examples shown in Figure~\ref{fig:ablation_pose}.

\subsubsection{Automatic Metrics}
To evaluate the performance of each model in terms of quality and diversity of their generated samples, we propose a Fr\'echet Pose Distance, FPD, computed as the Fr\`echet distance in the Vposer embedding space. Moreover, to measure the correctness of the generated poses, we compute the CLIP similarity score between the rendered avatars corresponding to each generated pose and their textual descriptions. As reported in Table~\ref{tab:pose_gen_eval}, our method outperforms the baseline MotionCLIP, AvatarCLIP, and MDM models significantly. Although TEMOS shows slightly better results in terms of FPD and CLIPSIM scores, it performs very poorly in our human evaluations. This could be due to overfitting to the mocap dataset and not generalizing to in-the-wild poses.  

\input{Tables/pose_generation_auto_eval.tex}

\subsubsection{Ablation Studies}
We perform another human evaluation to understand the impact of different continuous pose representations during training: 32-dimensional VPoser embedding for the body pose, 6D continuous vectors for all joints, and 6D continuous vectors for all joints regularized by the pre-trained VPoser, as explained in~\ref{sec:vposer}.  Our results summarized in Table~\ref{tab:ablation_data} confirm the benefit of VPoser representations either in the training of the diffusion model or used as a post regularizer (the last two rows in the table). We also compare the performance of our model when trained on the ITPP data vs. trained on the modified HumanML3D dataset for pose generation. This study again confirms the impact of our large-scale ITPP dataset compared with the limited existing MOCAP datasets by providing a wide range of human actions and training more generalizable motion generation models.

\input{Tables/ablation_table.tex}

\subsection{Evaluation of Personalized Image Generation}
We compare our approach for avatar-personalized image generation with a simple outpainting baseline which is fine-tuned from our pre-trained text-to-image model. For this evaluation, we provide as input the posed avatars and their corresponding mask. We also compare against this inpainting/outpainting model which has been fine-tuned on our ITPP dataset to decouple the value of the dataset versus our architecture modifications (termed ``fine-tuned'').

\begin{figure*}
  \centering \includegraphics[width=\textwidth]{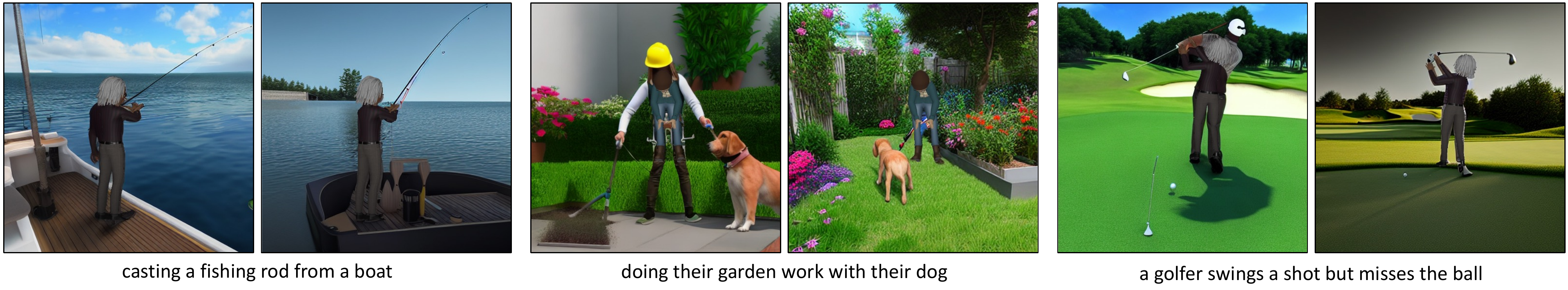}
  \caption{Qualitative comparison between the inpainting/outpainting baseline and our image generation model. For each caption, the baseline generation is on the left and ours is on the right. Captions are prefixed by ``A person (is) '' except the rightmost which has no prefix. Notice how the lack of human body priors in the baseline leads it to frequently hallucinate new limbs (first two examples) or misinterpret the human body mask (last example), resulting in poor avatar-scene composition. The same rendered avatar from our text-to-3D pose model is provided to both models.}
  \label{fig:image_gen_comp}
\end{figure*}
\vspace{-1mm}

We use the same test set of human evaluation prompts described in ~\ref{sec:human_eval_set}, and use poses generated from our text-to-3D pose generation model. We report both human evaluation and automatic metrics. For automatic metrics, we compute the mean CLIP similarity between the generated image and the input text prompt. For human evaluation, we report two metrics. To measure how well the avatar appearance corresponds to the ground truth avatar, we present raters with the input avatar as well as a generated image from our model and the baseline and ask which image is more faithful to the avatar's appearance. Note that while we blend the avatar's appearance back onto the generated image, there may still be hallucinated limbs or articles of clothing. To measure how well the avatar-scene interaction follows the text, we present raters with the input text as well as a generated image from our model and the baseline and ask which image better matches the text. We present each generation to 5 raters and report the majority vote. All human evaluation metrics are reported with respect to the inpainting/outpainting baseline.
\input{Tables/image_generation_human_eval.tex}

As we can see in Table~\ref{tab:image_gen_eval}, our method outperforms the baseline in both automatic and human evaluation metrics. We also show qualitative comparisons in Figure~\ref{fig:image_gen_comp}. In particular, our method makes significant improvements in faithfulness to the avatar's appearance. This makes sense because the inpainting/outpainting baseline model was trained on random masks, not human masks. As such, we find that it often hallucinates new limbs and articles of clothing, disturbing the avatar's identity (see Fig.~\ref{fig:image_gen_comp} for examples). Both the fine-tuning on human body masks and the architecture modifications improve faithfulness to the avatar appearance. In particular, we believe that the additional conditioning of the image generator on the gray body render improves the model's understanding of the human body orientation and limb positioning to reduce hallucination of new body parts. 

\section{Limitations \& Future Work}
\begin{figure}
  \centering \includegraphics[width=0.5\textwidth]{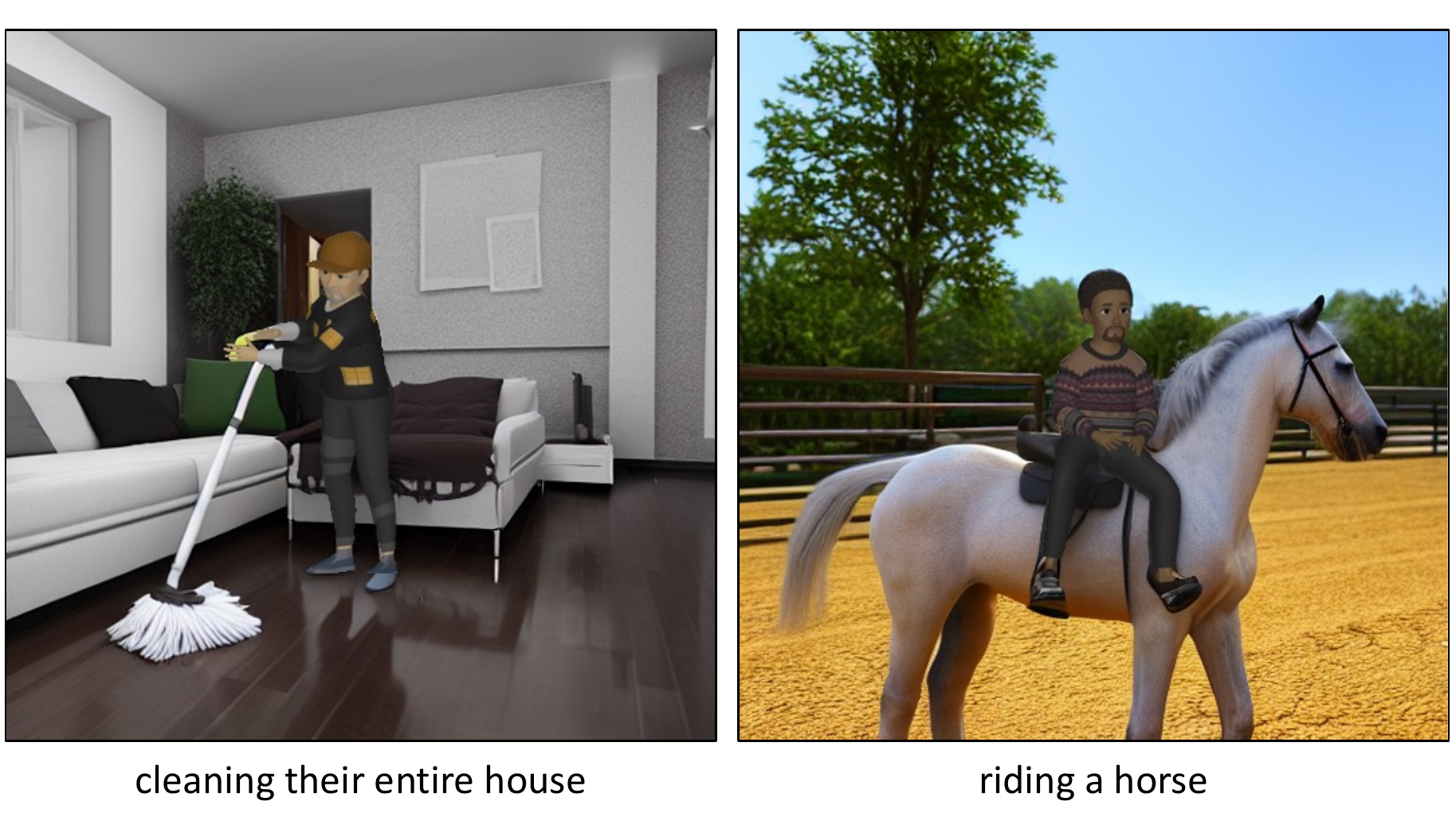}
  \caption{\textbf{Limitations:} In some cases, the lack of hand pose modeling limits our ability to generate fine-grained hand-object interactions (left). Other times, our model's inability to generate occlusions over the avatar's body results in unrealistic compositions (right).}
  \label{fig:limitations}
\end{figure}

There are a number of limitations of our approach which we leave for future work. For one, we only model human body pose in our text-to-3D pose diffusion model. This means that we are unable to control hand poses and facial expressions via text. Without modeling hand poses, we are limited in the granularity of avatar-object interactions (e.g., Figure~\ref{fig:limitations} left). In the future, we should be able to add hand and face pose parameters as additional targets for the text-to-3D pose diffusion model to enable an even greater level of expression through avatars. There are also limitations to our approach of treating personalized image generation as an outpainting task. Namely, we are unable to generate occlusions over parts of the avatar's body and the location of the avatar in the scene must be fixed before image generation. The inability to handle occlusions is particularly noticeable with prompts like ``A person is horseback riding'' where depending on the camera angle, one of the avatar's legs should be occluded (e.g., Figure~\ref{fig:limitations} right). 

\vspace{-2mm}

\section{Conclusions}
Motivated to facilitate self-expression of millions of users, we propose a pipeline for zero-shot personalization of image generation through avatars. In order to enable zero-shot personalization while remaining faithful to the avatar's identity, we first re-pose and render a user's avatar before generating a personalized image. Our proposed novel text-to-3D pose model beats existing baselines because of a new auto-generated dataset of 35M pairs of human poses and their text descriptions extracted from large-scale image datasets. In fact, our experiments show that this dataset produces better results than adding human motion capture datasets like AMASS in our pipeline. As such, this work paves the way to leverage large-scale image and video datasets for learning human 3D pose parameters.

\begin{figure*}
  \centering \includegraphics[width=0.9\textwidth]{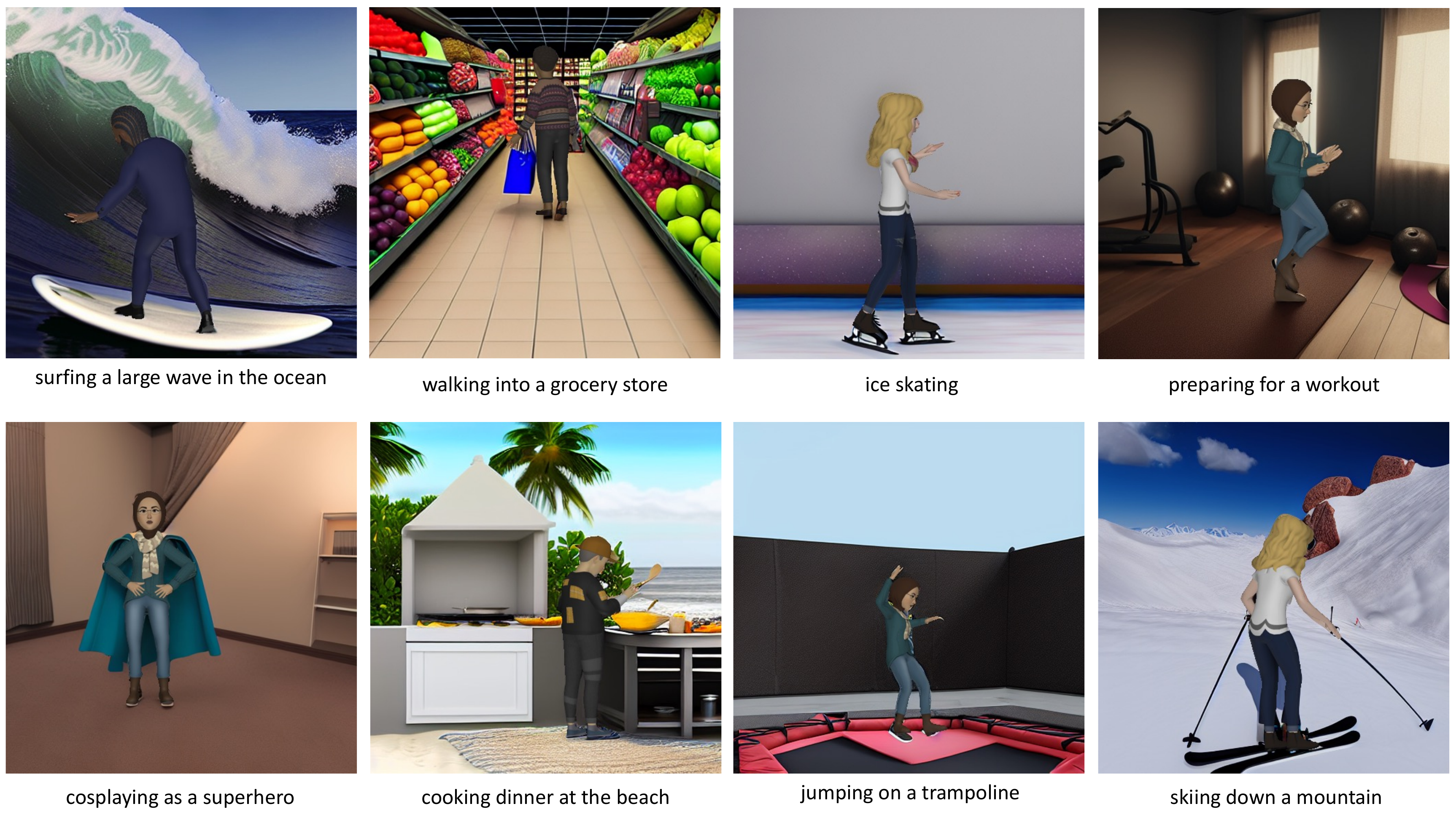}
  \caption{Additional image samples from PAS.}
  \label{fig:additional_samples}
\end{figure*}

\newpage
{\small
\bibliographystyle{ieee_fullname}
\bibliography{sample-bibliography}
}

\end{document}

%% file: Tables/pose_generation_auto_eval.tex
\begin{table}[t]
    \begin{center}
        \setlength{\tabcolsep}{2pt}
            \begin{tabular}{lccc}
                \toprule
                \textbf{Method} & 
                Human Eval $\uparrow$ &
                FPD $\downarrow$ & 
                CLIPSIM $\uparrow$ \\
                \midrule
                TEMOS \cite{petrovich2022temos} &
                73.7 &
                4.6 &
                0.17 \\
                MotionCLIP \cite{tevet2022motionclip} &
                59.0 &
                12.06 &
                0.145 \\
                AvatarCLIP \cite{hong2022avatarclip}  &
                60.7 &
                13.28 &
                0.162 \\
                MDM \cite{tevet2022mdm}  &
                60.0 &
                11.16 &
                0.15 \\
                 \bottomrule
                PAS (6D+VPoser) &
                - &
                6.25 &
                0.16 \\
                PAS (VPoser) &
                - &
                7.27 &
                0.164 \\
                \bottomrule
            \end{tabular}
    \caption{\textbf{Text-to-3D Pose Generation Evaluation:} We evaluate our method, PAS, against text-to-motion generation baselines. We report human evaluation scores where we compare each baseline against our model in terms of the correctness of each generated pose given a text prompt. The results are reported as the percentage of the majority vote raters that preferred our method to each baseline on our curated 390 prompts set. FPD measures overall quality and diversity of the samples, and CLIPSIM measures pose-text faithfulness; Both evaluated on the HumanML3D test set. Our model is trained on our ITPP dataset and HumanML3D train set.}
    \vspace{-2em}
    \label{tab:pose_gen_eval}
    \end{center}
\end{table}

%% file: Tables/ablation_table.tex
\begin{table}[t]
    \begin{center}
        \setlength{\tabcolsep}{2pt}
            \begin{tabular}{lccc}
                \toprule
                \textbf{Method} & 
                Training data &
                Human Eval $\uparrow$ &
                CLIPSIM  $\uparrow$ \\
                \midrule
                6D+VPoser  &
                HumanML3D &
                69.8 &
                0.149 \\
                6D  &
                ITPP &
                56.6 &
                0.146 \\
                VPoser &
                ITPP &
                51.5 &
                0.150 \\
                6D+VPoser &
                ITPP &
                - &
                0.149 \\

                \bottomrule
            \end{tabular}
    \caption{\textbf{Ablation study of different pose representations and training datasets}: HumanML3D dataset was modified for single frame pose generation. The human evaluation text faithfulness is reported as the percentage of the majority vote raters that preferred the last line setting over the others on our curated 390 prompts set, i.e., any value above 50\% means 6D+Vposer trained on our ITPP data was favored. For CLIP similarity, we rendered SMPL avatars using the 3D pose predicted by our model. }
    \vspace{-2em}
    \label{tab:ablation_data}
    \end{center}
\end{table}

%% file: Tables/image_generation_human_eval.tex
\begin{table}[t]
    \begin{center}
        \setlength{\tabcolsep}{2pt}
            \begin{tabular}{lccc}
                \toprule
                \textbf{Method} & 
                Avatar $\uparrow$ & 
                Text $\uparrow$ &
                CLIPSIM  $\uparrow$
                \\
                &
                match & 
                match &
                \\
                \midrule
                 Baseline &
                - &
                - &
                0.239
                \\

                Fine-tuned &
                0.56 &
                0.48 &
                0.239
                \\

                Ours  &
                \textbf{0.63} &
                \textbf{0.52} &
                \textbf{0.242}
                \\

                \bottomrule
            \end{tabular}
    \caption{\textbf{Personalized Image Generation Evaluation:} We evaluated our method against an inpainting/outpainting baseline. We report human evaluation (avatar and text faithfulness) and automatic metrics (CLIP similarity). The results for the human evaluation are reported as the percentage of the majority vote raters that preferred our method over the baseline, i.e., any value above 50\% means our method was favored. The ``fine-tuned'' method is the inpainting/outpainting model fine-tuned on our ITPP dataset, and isolates the value of the architecture modifications we made.}
    \vspace{-2em}
    \label{tab:image_gen_eval}
    \end{center}
\end{table}

%% file: arxiv.bbl
\begin{thebibliography}{10}\itemsep=-1pt

\bibitem{aberman2019deep}
Kfir Aberman, Mingyi Shi, Jing Liao, Dani Lischinski, Baoquan Chen, and Daniel
  Cohen-Or.
\newblock Deep video-based performance cloning.
\newblock In {\em Computer Graphics Forum}, volume~38, pages 219--233. Wiley
  Online Library, 2019.

\bibitem{ahuja2019jl2p}
Chaitanya Ahuja and Louis-Philippe Morency.
\newblock Language2pose: Natural language grounded pose forecasting.
\newblock {\em IEEE International Conference on 3D Vision (3DV)}, pages
  719--728, 2019.

\bibitem{azadi2020compositional}
Samaneh Azadi, Deepak Pathak, Sayna Ebrahimi, and Trevor Darrell.
\newblock Compositional gan: Learning image-conditional binary composition.
\newblock {\em International Journal of Computer Vision}, 128(10):2570--2585,
  2020.

\bibitem{briq2021}
Rania Briq, Pratika Kochar, , and Juergen Gall.
\newblock Towards better adversarial synthesis of human images from text.
\newblock {\em arXiv preprint arXiv:2107.01869}, 2021.

\bibitem{cervantes2022}
Pablo Cervantes, Yusuke Sekikawa, Ikuro Sato, and Koichi Shinoda.
\newblock Implicit neural representations for variable length human motion
  generation.
\newblock {\em European Conference on Computer Vision (ECCV)}, 2022.

\bibitem{chan2019everybody}
Caroline Chan, Shiry Ginosar, Tinghui Zhou, and Alexei~A Efros.
\newblock Everybody dance now.
\newblock In {\em Proceedings of the IEEE/CVF international conference on
  computer vision}, pages 5933--5942, 2019.

\bibitem{diff_beat_gans}
Prafulla Dhariwal and Alexander Nichol.
\newblock Diffusion models beat gans on image synthesis.
\newblock {\em Advances in Neural Information Processing Systems (NeurIPS)},
  34, 2021.

\bibitem{make_a_scene}
Oran Gafni, Adam Polyak, Oron Ashual, Shelly Sheynin, Devi Parikh, and Yaniv
  Taigman.
\newblock Make-a-scene: Scene-based text-to-image generation with human priors.
\newblock {\em European Conference on Computer Vision (ECCV)}, 2022.

\bibitem{ghosh2021}
Anindita Ghosh, Noshaba Cheema, Cennet Oguz, Christian Theobalt, and Philipp
  Slusallek.
\newblock Synthesis of compositional animations from textual descriptions.
\newblock {\em Proceedings of the IEEE/CVF International Conference on Computer
  Vision (ICCV)}, pages 1396--1406, October 2021.

\bibitem{goodfellow2014}
Ian Goodfellow, Jean Pouget-Abadie, Mehdi Mirza, Bing Xu, David Warde-Farley,
  Sherjil Ozair, Aaron Courville, and Yoshua Bengio.
\newblock Generative adversarial nets.
\newblock In {\em Advances in Neural Information Processing Systems}, 2014.

\bibitem{grigorev2019coordinate}
Artur Grigorev, Artem Sevastopolsky, Alexander Vakhitov, and Victor Lempitsky.
\newblock Coordinate-based texture inpainting for pose-guided human image
  generation.
\newblock In {\em Proceedings of the IEEE/CVF Conference on Computer Vision and
  Pattern Recognition}, pages 12135--12144, 2019.

\bibitem{guo2022t2m}
Chuan Guo, Shihao Zou, Xinxin Zuo, Sen Wang, Wei Ji, Xingyu Li, and Li Cheng.
\newblock Generating diverse and natural 3d human motions from text.
\newblock {\em Proceedings of the IEEE/CVF Conference on Computer Vision and
  Pattern Recognition (CVPR)}, pages 5152--5161, June 2022.

\bibitem{HumanML3d}
Chuan Guo, Shihao Zou, Xinxin Zuo, Sen Wang, Wei Ji, Xingyu Li, and Li Cheng.
\newblock Generating diverse and natural 3d human motions from text.
\newblock In {\em Proceedings of the IEEE/CVF Conference on Computer Vision and
  Pattern Recognition (CVPR)}, pages 5152--5161, June 2022.

\bibitem{guo2020a2m}
Chuan Guo, Xinxin Zuo, Sen Wang, Shihao Zou, Qingyao Sun, Annan Deng, Minglun
  Gong, and Li Cheng.
\newblock Action2motion: Conditioned generation of 3d human motions.
\newblock {\em Proceedings of the 28th ACM International Conference on
  Multimedia}, pages 2021--2029, 2020.

\bibitem{hertz2022prompt}
Amir Hertz, Ron Mokady, Jay Tenenbaum, Kfir Aberman, Yael Pritch, and Daniel
  Cohen-Or.
\newblock Prompt-to-prompt image editing with cross attention control.
\newblock {\em arXiv preprint arXiv:2208.01626}, 2022.

\bibitem{ddpms}
Jonathan Ho, Ajay Jain, and Pieter Abbeel.
\newblock Denoising diffusion probabilistic models.
\newblock {\em Advances in Neural Information Processing Systems (NeurIPS)},
  33, 2020.

\bibitem{cfg}
Jonathan Ho and Tim Salimans.
\newblock Classifier-free diffusion guidance.
\newblock {\em arXiv preprint arXiv:2207.12598}, 2022.

\bibitem{hong2022avatarclip}
Fangzhou Hong, Mingyuan Zhang, Liang Pan, Zhongang Cai, Lei Yang, and Ziwei
  Liu.
\newblock Avatarclip: Zero-shot text-driven generation and animation of 3d
  avatars.
\newblock {\em ACM Transactions on Graphics (TOG)}, 41(4):1--19, 2022.

\bibitem{kappel2021high}
Moritz Kappel, Vladislav Golyanik, Mohamed Elgharib, Jann-Ole Henningson,
  Hans-Peter Seidel, Susana Castillo, Christian Theobalt, and Marcus Magnor.
\newblock High-fidelity neural human motion transfer from monocular video.
\newblock In {\em Proceedings of the IEEE/CVF conference on computer vision and
  pattern recognition}, pages 1541--1550, 2021.

\bibitem{kawar2022imagic}
Bahjat Kawar, Shiran Zada, Oran Lang, Omer Tov, Huiwen Chang, Tali Dekel, Inbar
  Mosseri, and Michal Irani.
\newblock Imagic: Text-based real image editing with diffusion models.
\newblock {\em arXiv preprint arXiv:2210.09276}, 2022.

\bibitem{lee2019}
Hsin-Ying Lee, Xiaodong Yang, Ming-Yu Liu, Ting-Chun Wang, Yu-Ding Lu,
  Ming-Hsuan Yang, and Jan Kautz.
\newblock Dancing to music.
\newblock {\em Advances in Neural Information Processing Systems (NeurIPS)},
  32, 2019.

\bibitem{li2021}
Ruilong Li, Shan Yang, David~A. Ross, and Angjoo Kanazawa.
\newblock Ai choreographer: Music conditioned 3d dance generation with aist++.
\newblock {\em Proceedings of the IEEE/CVF International Conference on Computer
  Vision (ICCV)}, 2021.

\bibitem{liu2020neural}
Lingjie Liu, Weipeng Xu, Marc Habermann, Michael Zollh{\"o}fer, Florian
  Bernard, Hyeongwoo Kim, Wenping Wang, and Christian Theobalt.
\newblock Neural human video rendering by learning dynamic textures and
  rendering-to-video translation.
\newblock {\em arXiv preprint arXiv:2001.04947}, 2020.

\bibitem{liu2023more}
Xihui Liu, Dong~Huk Park, Samaneh Azadi, Gong Zhang, Arman Chopikyan, Yuxiao
  Hu, Humphrey Shi, Anna Rohrbach, and Trevor Darrell.
\newblock More control for free! image synthesis with semantic diffusion
  guidance.
\newblock In {\em Proceedings of the IEEE/CVF Winter Conference on Applications
  of Computer Vision}, pages 289--299, 2023.

\bibitem{ma2017pose}
Liqian Ma, Xu Jia, Qianru Sun, Bernt Schiele, Tinne Tuytelaars, and Luc
  Van~Gool.
\newblock Pose guided person image generation.
\newblock {\em Advances in neural information processing systems}, 30, 2017.

\bibitem{ma2018disentangled}
Liqian Ma, Qianru Sun, Stamatios Georgoulis, Luc Van~Gool, Bernt Schiele, and
  Mario Fritz.
\newblock Disentangled person image generation.
\newblock In {\em Proceedings of the IEEE Conference on Computer Vision and
  Pattern Recognition}, pages 99--108, 2018.

\bibitem{julieta2017}
Julieta Martinez, Michael~J. Black, and Javier Romero.
\newblock On human motion prediction using recurrent neural networks.
\newblock {\em Proceedings of the IEEE/CVF Conference on Computer Vision and
  Pattern Recognition (CVPR)}, 2017.

\bibitem{meshry2019neural}
Moustafa Meshry, Dan~B Goldman, Sameh Khamis, Hugues Hoppe, Rohit Pandey, Noah
  Snavely, and Ricardo Martin-Brualla.
\newblock Neural rerendering in the wild.
\newblock In {\em Proceedings of the IEEE/CVF Conference on Computer Vision and
  Pattern Recognition}, pages 6878--6887, 2019.

\bibitem{Mirza2014ConditionalGA}
Mehdi Mirza and Simon Osindero.
\newblock Conditional generative adversarial nets.
\newblock {\em ArXiv}, abs/1411.1784, 2014.

\bibitem{nichol2021glide}
Alex Nichol, Prafulla Dhariwal, Aditya Ramesh, Pranav Shyam, Pamela Mishkin,
  Bob McGrew, Ilya Sutskever, and Mark Chen.
\newblock Glide: Towards photorealistic image generation and editing with
  text-guided diffusion models.
\newblock {\em arXiv preprint arXiv:2112.10741}, 2021.

\bibitem{GLIDE2022}
Alexander~Quinn Nichol, Prafulla Dhariwal, Aditya Ramesh, Pranav Shyam, Pamela
  Mishkin, Bob McGrew, Ilya Sutskever, and Mark Chen.
\newblock {GLIDE:} towards photorealistic image generation and editing with
  text-guided diffusion models.
\newblock In {\em International Conference on Machine Learning}, 2022.

\bibitem{SMPL-X:2019}
Georgios Pavlakos, Vasileios Choutas, Nima Ghorbani, Timo Bolkart, Ahmed A.~A.
  Osman, Dimitrios Tzionas, and Michael~J. Black.
\newblock Expressive body capture: 3d hands, face, and body from a single
  image.
\newblock In {\em Proceedings IEEE Conf. on Computer Vision and Pattern
  Recognition (CVPR)}, 2019.

\bibitem{petrovich2021actor}
Mathis Petrovich, Michael~J. Black, and G{\"u}l Varol.
\newblock Action-conditioned 3{D} human motion synthesis with transformer
  {VAE}.
\newblock {\em Proceedings of the IEEE/CVF International Conference on Computer
  Vision (ICCV)}, pages 10985--10995, October 2021.

\bibitem{petrovich2022temos}
Mathis Petrovich, Michael~J. Black, and Gül Varol.
\newblock Temos: Generating diverse human motions from textual descriptions.
\newblock {\em European Conference on Computer Vision (ECCV)}, 2022.

\bibitem{ramesh2022dalle2}
Aditya Ramesh, Prafulla Dhariwal, Alex Nichol, Casey Chu, and Mark Chen.
\newblock Hierarchical text-conditional image generation with clip latents.
\newblock {\em arXiv preprint arXiv:2204.06125}, 2022.

\bibitem{Ramesh2022HierarchicalTI}
Aditya Ramesh, Prafulla Dhariwal, Alex Nichol, Casey Chu, and Mark Chen.
\newblock Hierarchical text-conditional image generation with clip latents.
\newblock {\em ArXiv}, abs/2204.06125, 2022.

\bibitem{DALLE2021}
Aditya Ramesh, Mikhail Pavlov, Gabriel Goh, Scott Gray, Chelsea Voss, Alec
  Radford, Mark Chen, and Ilya Sutskever.
\newblock Zero-shot text-to-image generation.
\newblock In {\em Proceedings of the 38th International Conference on Machine
  Learning}, 2021.

\bibitem{reed16GANText}
Scott Reed, Zeynep Akata, Xinchen Yan, Lajanugen Logeswaran, Bernt Schiele, and
  Honglak Lee.
\newblock Generative adversarial text to image synthesis.
\newblock In {\em Proceedings of The 33rd International Conference on Machine
  Learning}, 2016.

\bibitem{roy2022}
Prasun Roy, Subhankar Ghosh, Saumik Bhattacharya, Umapada Pal, and Michael
  Blumenstein.
\newblock Tips: Text-induced pose synthesis.
\newblock {\em European Conference on Computer Vision (ECCV)}, 2022.

\bibitem{ruiz2019}
Alejandro~Hernandez Ruiz, Juergen Gall, and Francesc Moreno-Noguer.
\newblock Human motion prediction via spatio-temporal inpainting.
\newblock {\em Proceedings of the IEEE/CVF International Conference on Computer
  Vision (ICCV)}, 2019.

\bibitem{ruiz2022dreambooth}
Nataniel Ruiz, Yuanzhen Li, Varun Jampani, Yael Pritch, Michael Rubinstein, and
  Kfir Aberman.
\newblock Dreambooth: Fine tuning text-to-image diffusion models for
  subject-driven generation.
\newblock {\em arXiv preprint arXiv:2208.12242}, 2022.

\bibitem{imagen}
Chitwan Saharia, William Chan, Saurabh Saxena, Lala Li, Jay Whang, Emily
  Denton, Seyed Kamyar~Seyed Ghasemipour, Burcu~Karagol Ayan, S.~Sara Mahdavi,
  Rapha~Gontijo Lopes, Tim Salimans, Jonathan Ho, David~J Fleet, and Mohammad
  Norouzi.
\newblock Style and pose control for image synthesis of humans from a single
  monocular view.
\newblock {\em arXiv preprint arXiv:2205.11487}, 2022.

\bibitem{sarkar2021style}
Kripasindhu Sarkar, Vladislav Golyanik, Lingjie Liu, and Christian Theobalt.
\newblock Style and pose control for image synthesis of humans from a single
  monocular view.
\newblock {\em arXiv preprint arXiv:2102.11263}, 2021.

\bibitem{stylegan-T2023}
Axel Sauer, Tero Karras, Samuli Laine, Andreas Geiger, and Timo Aila.
\newblock Stylegan-t: Unlocking the power of gans for fast large-scale
  text-to-image synthesis.
\newblock volume abs/2301.09515, 2023.

\bibitem{stylegan-xl2022}
Axel Sauer, Katja Schwarz, and Andreas Geiger.
\newblock Stylegan-xl: Scaling stylegan to large diverse datasets.
\newblock In {\em ACM SIGGRAPH 2022 Conference Proceedings}, 2022.

\bibitem{tevet2022motionclip}
Guy Tevet, Brian Gordon, Amir Hertz, Amit~H. Bermano, and Daniel Cohen-Or.
\newblock Motionclip: Exposing human motion generation to clip space.
\newblock {\em European Conference on Computer Vision (ECCV)}, 2022.

\bibitem{tevet2022mdm}
Guy Tevet, Sigal Raab, Brian Gordon, Yonatan Shafir, Amit~H Bermano, and Daniel
  Cohen-Or.
\newblock Human motion diffusion model.
\newblock {\em arXiv preprint arXiv:2209.14916}, 2022.

\bibitem{wu2019detectron2}
Yuxin Wu, Alexander Kirillov, Francisco Massa, Wan-Yen Lo, and Ross Girshick.
\newblock Detectron2.
\newblock \url{https://github.com/facebookresearch/detectron2}, 2019.

\bibitem{yan2019}
Sijie Yan, Zhizhong Li, Yuanjun Xiong, Huahan Yan, and Dahua Lin.
\newblock Convolutional sequence generation for skeleton-based action
  synthesis.
\newblock {\em Proceedings of the IEEE/CVF International Conference on Computer
  Vision (ICCV)}, pages 4393--4401, 2019.

\bibitem{Parti}
Jiahui Yu, Yuanzhong Xu, Jing~Yu Koh, Thang Luong, Gunjan Baid, Zirui Wang,
  Vijay Vasudevan, Alexander Ku, Yinfei Yang, Burcu~Karagol Ayan, Ben
  Hutchinson, Wei Han, Zarana Parekh, Xin Li, Han Zhang, Jason Baldridge, and
  Yonghui Wu.
\newblock Human motion diffusion model.
\newblock {\em arXiv preprint arXiv:2206.10789}, 2022.

\bibitem{pymafx}
Hongwen Zhang, Yating Tian, Yuxiang Zhang, Mengcheng Li, Liang An, Zhenan Sun,
  and Yebin Liu.
\newblock Pymaf-x: Towards well-aligned full-body model regression from
  monocular images.
\newblock {\em arXiv preprint arXiv:2207.06400}, 2022.

\bibitem{zhang2022md}
Mingyuan Zhang, Zhongang Cai, Liang Pan, Fangzhou Hong, Xinying Guo, Lei Yang,
  and Ziwei Liu.
\newblock Motiondiffuse: Text-driven human motion generation with diffusion
  model.
\newblock {\em arXiv preprint arXiv:2208.15001}, 2022.

\bibitem{zhang2021}
Yifei Zhang, Rania Briq, Julian Tanke, and Juergen Gall.
\newblock Adversarial synthesis of human pose from text.
\newblock {\em 42nd DAGM German Conference of Pattern Recognition}, pages
  145--158, 2021.

\bibitem{zhao2020}
Rui Zhao, Hui Su, and Qiang Ji.
\newblock Bayesian adversarial human motion synthesis.
\newblock {\em Proceedings of the IEEE/CVF Conference on Computer Vision and
  Pattern Recognition (CVPR)}, pages 6224--6233, 2020.

\bibitem{zhou2019pose}
Xingran Zhou, Siyu Huang, Bin Li, Yingming Li, Jiachen Li, and Zhongfei Zhang.
\newblock Text guided person image synthesis.
\newblock {\em Proceedings IEEE Conf. on Computer Vision and Pattern
  Recognition (CVPR)}, 2019.

\bibitem{zhou2019continuity}
Yi Zhou, Connelly Barnes, Jingwan Lu, Jimei Yang, and Hao Li.
\newblock On the continuity of rotation representations in neural networks.
\newblock In {\em Proceedings of the IEEE/CVF Conference on Computer Vision and
  Pattern Recognition}, pages 5745--5753, 2019.

\bibitem{zhu2019progressive}
Zhen Zhu, Tengteng Huang, Baoguang Shi, Miao Yu, Bofei Wang, and Xiang Bai.
\newblock Progressive pose attention transfer for person image generation.
\newblock In {\em Proceedings of the IEEE/CVF Conference on Computer Vision and
  Pattern Recognition}, pages 2347--2356, 2019.

\end{thebibliography}
